# Comparative Analysis of N-gram Text Representation on Igbo Text Document Similarity

Ifeanyi-Reuben Nkechi J.
Department of Computer Science,
Rhema University, Nigeria

Ugwu Chidiebere
Department of Computer Science
University of Port Harcourt, Nigeria

Nwachukwu E. O.
Department of Computer Science
University of Port Harcourt, Nigeria

## ABSTRACT
The improvement in Information Technology has encouraged the use of Igbo in the creation of text such as resources and news articles online. Text similarity is of great importance in any text-based applications. This paper presents a comparative analysis of n-gram text representation on Igbo text document similarity. It adopted Euclidean similarity measure to determine the similarities between Igbo text documents represented with two word-based n-gram text representation (unigram and bigram) models. The evaluation of the similarity measure is based on the adopted text representation models. The model is designed with Object-Oriented Methodology and implemented with Python programming language with tools from Natural Language Toolkits (NLTK). The result shows that unigram represented text has highest distance values whereas bigram has the lowest corresponding distance values. The lower the distance value, the more similar the two documents and better the quality of the model when used for a task that requires similarity measure. The similarity of two documents increases as the distance value moves down to zero (0). Ideally, the result analyzed revealed that Igbo text document similarity measured on bigram represented text gives accurate similarity result. This will give better, effective and accurate result when used for tasks such as text classification, clustering and ranking on Igbo text.

## 1. INTRODUCTION
The extraction and management of useful information that are hidden in the huge quantity of text documents together with its unstructured nature has generated big concern to IT professionals. It is difficult and uneconomical to explore the useful information present in the unstructured text documents of web pages, text content management system, news articles, and others [15]. Text document similarity measure plays a vital role in text-based intelligent tasks such as text classification, text clustering, plagiarism detection and information retrieval [4].

The advancement of Information Technology has encouraged the use of Igbo language in the creation of resources, articles and news reports online [1]. As the number of Igbo texts online and tasks that require computing similarity between Igbo text documents are increasing, there is need to have an effective model to measure the degree of closeness between these texts for good knowledge management.

This paper compares the effect of word-based n-gram text representation model on Igbo text document similarity. It calculates the degree of similarity between Igbo text documents for efficient and effective text classification, clustering and ranking. N-gram models compared are unigram and bigram. A unigram model is an n-gram of size 1 and represents text in single words. Bigram is an n-gram of n = 2; and represents text in sequence of two words. [10].

### 1.1 Igbo Language Structure
Igbo is one of the three main languages (Hausa, Yoruba and Igbo) in Nigeria. It is largely spoken by the people in the eastern part of Nigeria. Igbo language has many dialects. The standard Igbo is used formally for this work. The current Igbo orthography (Onwu Orthography, 1961) is based on the Standard Igbo. Orthography is a way of writing sentence or constructing grammar in a language. Standard Igbo has thirty-six (36) alphabets (a, b, ch, d, e, f, g, gb, gh, gw, h, i, ị, j, k, kw, kp, l, m, n, nw, ny, ṅ, o, ọ, p, r, s, sh, t, u, ụ,v, w, y, z) [6].

Igbo language is agglutinative in nature; its words are formed by stringing up different words. Igbo vocabulary is made up of a large number of compound words. The individual meaning of words in a phrase or compound word does not entail the context it is being used for [1].

### 1.2 Overview of Similarity Measure
A similarity measure quantifies the similarity between two documents that reflects the degree of closeness or separation of the documents [5].

Sapna C., et al. [3] presented four similarity measure techniques as follows:

i. Cosine similarity measure: Cosine similarity measure uses the cosine of angle between two vectors. It measures similarity by computing the cosine angle between doc1 and doc2 as follows:

Cosine Similarity Measure (doc1, doc2) =

$$\frac{doc1.doc2}{\sqrt{(doc1.doc1)}\sqrt{(doc2.doc2)}}$$

ii. Jaccard similarity measure: Jaccard similarity measure is computed by finding the quotient of the size of the intersection of sample document set and the size of the union of the document set. This is computed as follows:

Jaccard Similarity Measure (doc1, doc2) =

$$\frac{doc1.doc2}{doc1.doc1 + doc2.doc2 - doc1.doc2}$$

iii. Dice Coefficient Similarity Measure: Dice coefficient similarity is described as the product of two and the number of terms which are common in the compared text documents and divided by the total number of features present in both documents [2]. This is given as follows:

Dice Coefficient Similarity Measure (doc, doc2) =

$$\frac{2 doc1.doc2}{doc1.doc1 + doc2.doc2}$$

iv. Euclidean Distance Similarity Measure: This uses Euclidean distance metric and its similarity value is obtained by computing the square root of the sum of squared differences between the text document features or attributes.

The paper adopted Euclidean distance metric to determine the similarities between the Igbo text documents.

## 2. RELATED WORKS

Some papers related to the work were studied and are discussed below:

In 2017, Ifeanyi-Reuben, N.J. et al [1] presented a paper on the analysis and representation of Igbo text document for a text-based intelligent system. The result obtained in their experiment showed that bigram and trigram n-gram model gives an ideal representation of Igbo text document for processing text analyses, considering the compounding nature of the language.

Vijaymeena, M.K. and Kavitha, K. [2] studied the existing approaches on text similarity and summarized them into three important approaches. These approaches are string-oriented, knowledge-oriented and corpus-oriented. String-oriented approach measures the similarity between text document using character and string series. Corpus-oriented approach measures the similarity between words based on the information obtained from large corpus. Knowledge-oriented, also known as the semantic similarity measure, determines the similarity between text documents based on the extent of similarity between the words and concepts. They recommended a combination of the approaches in a research to get hybrid similarity measures.

Mirza, R. M. and Losarwar, V. A. undertook a detailed work on similarity measure, cosine similarity and Euclidean distance. They evaluated the effects of the three measures on a real-world datasets and tested it on the text classification and clustering problems. The similarity measure gave optimal performance in their experiment [4].

Essam, S. H. [13] built similar thesaurus on Arabic language with full word and stemmed methods. The result was compared and it proved that similar thesaurus using stemmed method is more efficient than traditional full words method with higher levels of recall and precision.

## 3. MATERIALS AND METHODS

The architectural design of the model is given in Fig 1 and consists of four (4) basic modules: Igbo Text document collections, Text pre-processing, Text representation and Similarity measurement.

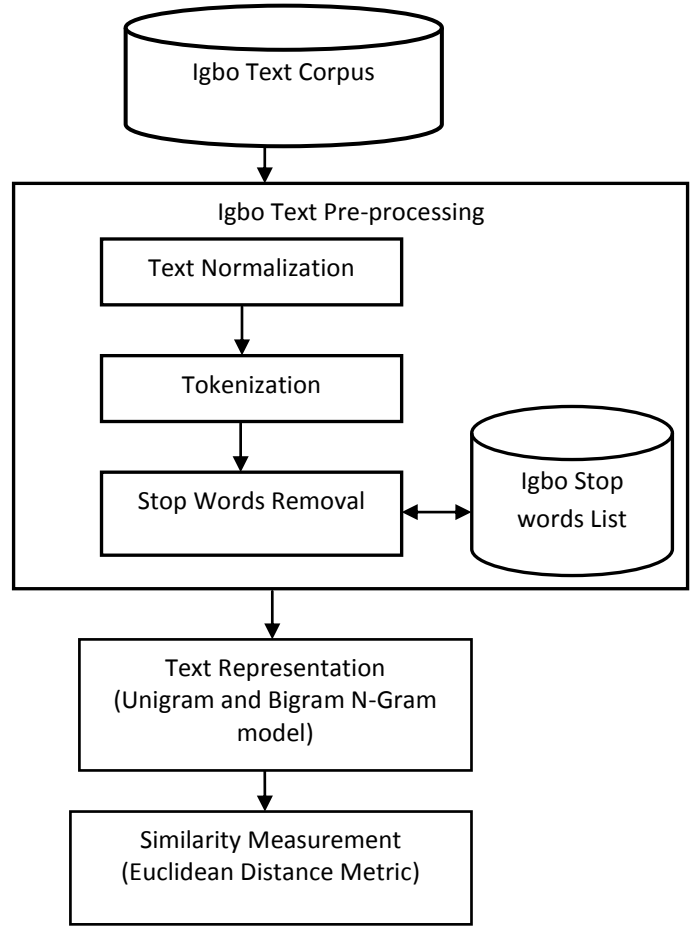

**Fig 1: Architecture of Igbo Text Similarity Measure**

### 3.1 Igbo Text Document Collections

The operation of any text-based system starts with the collection of text documents. The Unicode model was used for extracting and processing Igbo texts from file because it is one of the languages that employ non-ASCII character sets like English. Processing Igbo text needs UTF-8 encoding [1]. UTF-8 makes use of multiple bytes and represents complete collection of Unicode characters. This is achieved with the mechanisms of decoding and encoding as shown in Fig 2. Decoding translates text in files in a particular encoding like the Igbo text written with Igbo character sets into Unicode while encoding write Unicode to a file and convert it into an appropriate encoding [6]. The sources for the Igbo text documents collection of the paper are as follows:

i. Igbo Radio - Online News reports in Igbo language (www.igboradio.com).

ii. Rex Publications - Catholic Weekly Bulletin.

iii. Microsoft Igbo Language Software Localization Data

iv. Igbo Novels

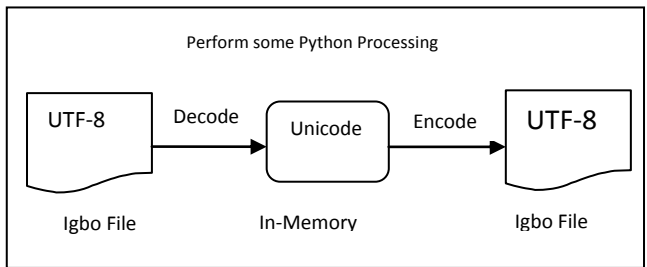

**Fig 2: Igbo Text Unicode Decoding and Encoding**

Kpaacharụ anya makana projektọ nkụziihe a achọghị okwu ntụghe, ndị ịchọghị ka ha hụ ga ahụ ihe-ngosi gi. Ọbụrụ na ịchọrọ iji projektọ nkụziihe a were rụọ ọrụ, pịkịnye "Jịkọọ". A na-akwụnye projektọ nkụziihe na kọmputa nkunaka iji mee ihe onyonyo. kọmputa nkunaka banyere na projektọ nkụziihe ọcha.

**Fig 3: Sample of Igbo Text Document**

## 3.2 Igbo Text Pre-processing

Text pre-processing is an essential task and an important step in any text-based tasks. This module transformed unstructured input of Igbo text into a more understandable and structured format ready for further processing [7]. The text pre-processing tasks in this work covers text normalization, Igbo text tokenization and Igbo stop-words removal. Tokenization analyzed and separated Igbo text into a sequence of discrete tokens (words). Stop-words removal removes the Igbo stop-words and Igbo language-specific functional words from the text. These are the most frequently used words in Igbo language that do not convey useful information. An Igbo stop-word list is created. The procedure for the text pre-processing task is detailed in algorithm 1.

**Algorithm 1:** Designed Igbo Text Pre-processing Algorithm.

Input: Igbo Text Document, Non-Igbo Standard Data/Character list.

Output: Pre-processed Igbo Text

Procedure:

1. Perform Igbo text normalization task.

a. Transform all text cases to lower case.

b. Remove diacritics (characters like ū, ù, and ú contain diacritics called tone marks).

c. Remove non-Igbo standard data/character:

d. For every word in the Text Document,

  i. IF the word is a digits (0, 1, 2, 3, 4, 5, 6, 7, 8, 9) or contains digits THEN the word is not useful, remove it.

  ii. If the word is a special character (:, ;, ?, !, ', (, ), {, }, +, &, [, ], <, >, /, @, ", !, *, =, ^, %, and others ) or contains special character, the word is non-Igbo, filter it out.

  iii. If the word is combined with hyphen like "nje-ozi", "na-aga", then remove hyphen and separate the words. If the word contains apostrophe like n'elu, n'ụlọ akwụkwọ, then remove the apostrophe and separate the words.

2. Perform Igbo Text Tokenization task:

  a. Create a TokenList.
  b. Increase the Token List if any.
  c. Separate characters or words, if the string matches any of the following: "ga-", "aga-", "n'", "na-", "ana-", "ọga-", "ịga-", "ọna-", "ịna-".
  d. Remove diacritics. This involves any non-zero length sequence of a–z, with grave accent (`), or acute accent (´),
  e. Any string separated with a whitespace is a token.
  f. Any single string enclosed in double or single quote" is a token.
  g. Any single string that ends with comma (,) or colon (:) or semi-colon (;) or exclamation mark (!) or question mark (?) or dot (.), is a token.

3. Perform Igbo Stop-words removal process:

  a. Read the stop-word file.
  b. Convert all loaded stop words to lower case.
  c. Read each word in the created Token List.
  d. For each word, w in Token List of the document

    i. Check if w(Token List) is in Language stop word list;
    ii. If Yes, remove w(Token List) from the Token List;
    iii. Decrease tokens count;
    iv. Move to the next w(Token List);
    v. If No, move to the next w(Token List);
    vi. Exit Iteration Loop;
    vii. Do the next task in the pre-processing process.

4. Remove any word with less than three (3) character length.

## 3.3 Text Representation

Text representation involves the selection of appropriate features to represent a document [8]. The approach in which text is represented has a big effect in the performance of any text-based applications [9]. It is strongly influenced by the language of the text.

N-gram model is adopted to represent Igbo text because of the compound nature of the language. The "N" spanned across 1 to 2, that is unigram and bigram. Unigram adopts the Bag-Of-Words (BOW) model and represents the Igbo text in single words. Bigram represents the Igbo text in sequence of two (2) words. The result of the two models on the similarity measure is analyzed to find the n-gram model that gives the optimal performance on the similarity measure of Igbo text documents.

Table 1 and Table 2 show the unigram and bigram representation of a sampled Igbo text document in Fig 3 after pre-processing.

**Table 1. Unigram Representation of Sample Igbo Text**

| S/No | Features | Frequency |
|---|---|---|
| 1 | achọghị | 1 |
| 2 | akwụnye | 1 |
| 3 | anya | 1 |
| 4 | banyere | 1 |
| 5 | ihe | 2 |
| 6 | iji | 2 |
| 7 | jikọọ | 1 |
| 8 | kpaacharụ | 1 |
| 9 | kọmputa | 2 |
| 10 | mee | 1 |
| 11 | ngosi | 1 |
| 12 | nkunaka | 2 |
| 13 | nkụziihe | 4 |
| 14 | ntụghe | 1 |
| 15 | okwu | 1 |
| 16 | onyonyo | 1 |
| 17 | prọjektọ | 4 |
| 18 | pịkịnye | 1 |
| 19 | rụọ | 1 |
| 20 | were | 1 |
| 21 | ịchọghị | 1 |
| 22 | ịchọrọ | 1 |
| 23 | ọbụrụ | 1 |
| 24 | ọcha | 1 |
| 25 | ọrụ | 1 |

**Table 2. Bigram Representation of Sample Igbo Text**

| S/No | Features | Frequency |
|---|---|---|
| 1 | prọjektọ nkụziihe | 4 |
| 2 | kọmputa nkunaka | 2 |
| 3 | kpaacharụ anya | 1 |
| 4 | anya prọjektọ | 1 |
| 5 | nkụziihe achọghị | 1 |
| 6 | achọghị okwu | 1 |
| 7 | okwu ntụghe | 1 |
| 8 | ntụghe ịchọghị | 1 |
| 9 | ịchọghị ihe | 1 |
| 10 | ihe ngosi | 1 |
| 11 | ngosi ọbụrụ | 1 |
| 12 | ọbụrụ ịchọrọ | 1 |
| 13 | ịchọrọ iji | 1 |
| 14 | iji prọjektọ | 1 |
| 15 | nkụziihe were | 1 |
| 16 | were rụọ | 1 |
| 17 | rụọ ọrụ | 1 |
| 18 | ọrụ pịkịnye | 1 |
| 19 | pịkịnye jikọọ | 1 |
| 20 | jikọọ akwụnye | 1 |
| 21 | akwụnye prọjektọ | 1 |
| 22 | nkụziihe kọmputa | 1 |
| 23 | nkunaka iji | 1 |
| 24 | iji mee | 1 |
| 25 | mee ihe | 1 |
| 26 | ihe onyonyo | 1 |
| 27 | onyonyo kọmputa | 1 |
| 28 | nkunaka banyere | 1 |
| 29 | banyere prọjektọ | 1 |
| 30 | nkụziihe ọcha | 1 |

### 3.4 Similarity Measurement

The distance metric is used to compute the similarities between various Igbo text documents. Algorithm 2 explains the procedures used in measuring the Igbo text similarity.

**Algorithm 2:** Measuring similarities in Igbo Text

**Procedure**: Find the similarity measure

**Input**: Igbo text documents.

Output: Similarity value/score.

1. Input Igbo text documents.
2. Perform text normalization.
3. Perform pre-processing.
4. Select relevant features.
5. For each vi in V and each wi in W do

    Calculate distance (vi,wi) based on distance measure

6. Determine the similarity of the texts based on the calculated distance d.

### 3.4.1 Euclidean Distance Metric

**Definition 1**: The Euclidean distance of points, a and b, is the length of the line segment connecting (ab). Given that $A = (a_1, a_2, ... a_n)$ and $B = (b_1, b_2, ...b_n)$ are two points in n-dimensional Euclidean space, then the distance (d) from A to B or B to A is given by the formula:

$$d^2(A,B) = d^2(B,A) = (a_1-b_1)^2 + (a_2-b_2)^2 + ... (a_n-b_n)^2$$

**Definition 2:** Assuming, Igbo text1 and Igbo text2 are two Igbo text documents in a corpus; and d(Igbo text1, Igbo text2) be the distance between Igbo text1 and Igbo text2, then:

i. d(Igbo text1, Igbo text2) >= 0; the distance between two documents cannot be zero (0);

ii. d(Igbo text1, Igbo text2) = 0 if and only if Igbo text1 = Igbo text2; the distance of two perfect similar documents is zero (0); the distance between document1 and document2 is the same distance between document2 and document1;

iii. d(Igbo text1, Igbo text2) = d(Igbo text2, Igbo text1)

Euclidean distance metric function measures distance between various points in a multidimensional data in real applications.

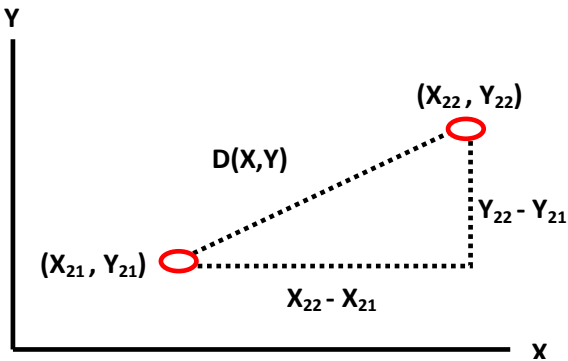

**Fig 4: Euclidean distance between x and y**

Fig. 4 illustrates the Euclidean distance between 2 points. Assuming the two points are $(X_{21}, Y_{21})$ and $(X_{22}, Y_{22})$ with the following features frequencies:

Ist Point: $X_{21} = 4$ and $Y_{21} = 2$

$2^{nd}$ Point: $X_{22} = 6$ and $Y_{22} = 5$

Then, the Euclidean distance between 2 points $(X_{21}, Y_{21})$ and $(X_{22}, Y_{22})$ is

$D(X,Y) = SQRT((X_{22}-X_{21})^2 + (Y_{22}-Y_{21})^2) = SQRT((6-4)^2 + (5-2)^2) = 3.32$. This means that the X and Y are 3.32 apart.

In the illustration of the Euclidean distance metric above, the X represents the first Igbo text document and Y represents the second Igbo text document. $X_{21}$, $X_{22}$, $Y_{21}$ and $Y_{22}$ are frequencies of features attributed to X and Y. Therefore, the distance between first Igbo text document and second Igbo text document is 3.32 (similarity value).

## 4. EXPERIMENTS

Total of eleven (11) Igbo text documents were used to test the model. This was done on Igbo text documents represented in unigram and bigram. Table 3, Table 4, and Table 5 display the summary of the results obtained. In Table 3 and Table 4, Igbo TextDoc1, Igbo TextDoc2, Igbo TextDoc3, Igbo TextDoc4, Igbo TextDoc5 and Igbo TextDoc6 represent column names while Igbo Text1, Igbo Text2, Igbo Text4 and Igbo Text5 are the row names. The distances between each document in row is computed across all documents in column.

**Table 3. Igbo Similarity Measure Result on Unigram Represented Text**

| Igbo Text | Igbo Text Doc1 | Igbo Text Doc2 | Igbo Text Doc3 | Igbo Text Doc4 | Igbo Text Doc5 | Igbo Text Doc6 |
|---|---|---|---|---|---|---|
| **Igbo Text1** | **6.78** | **4.36** | **6.40** | **5.48** | **7.07** | **7.28** |
| Igbo Text2 | 8.60 | 6.48 | 2.83 | 7.75 | 8.72 | 8.06 |
| Igbo Text4 | 0.00 | 7.28 | 10.20 | 6.86 | 10.82 | 10.68 |
| Igbo Text5 | 14.70 | 0.00 | 8.31 | 7.28 | 18.41 | 9.43 |
| Igbo Text6 | 9.80 | 8.00 | 0.00 | 9.06 | 9.90 | 9.49 |

**Table 4. Igbo Similarity Measure Result on Bigram Represented Text**

| Igbo Text | Igbo Text Doc1 | Igbo Text Doc2 | Igbo Text Doc3 | Igbo Text Doc4 | Igbo Text Doc5 | Igbo Text Doc6 |
|---|---|---|---|---|---|---|
| **Igbo Text1** | **5.00** | **6.00** | **4.47** | **3.61** | **7.48** | **5.00** |
| Igbo Text2 | 5.74 | 6.86 | 2.00 | 6.32 | 8.00 | 5.74 |
| Igbo Text4 | 0.00 | 7.81 | 6.71 | 7.21 | 9.00 | 7.07 |
| Igbo Text5 | 7.81 | 0.00 | 7.68 | 8.43 | 9.75 | 8.00 |
| Igbo Text6 | 6.71 | 7.68 | 0.00 | 7.21 | 8.72 | 6.71 |

**Table 5. Similarity Average Score on Unigram and Bigram Represented Igbo Text**

| Igbo Text | Unigram | Bigram |
|---|---|---|
| Igbo Text1 | 6.23 | 5.26 |
| Igbo Text2 | 7.07 | 5.78 |
| Igbo Text4 | 7.64 | 6.30 |
| Igbo Text5 | 9.69 | 6.95 |

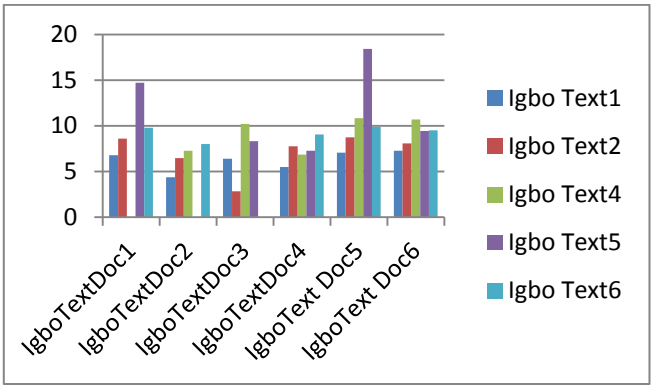

**Fig 5: Igbo Similarity Result on Unigram Text Chart**

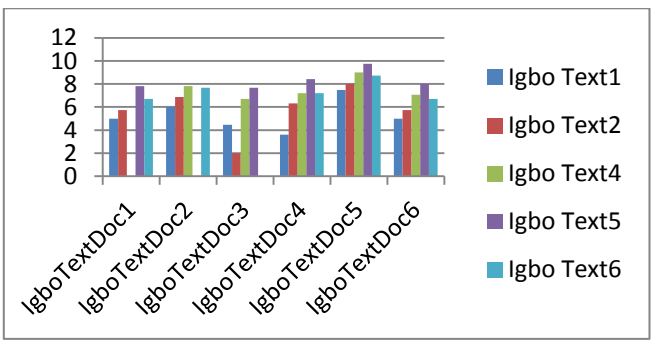

**Fig 6: Igbo Similarity Result on Bigram Text Chart**

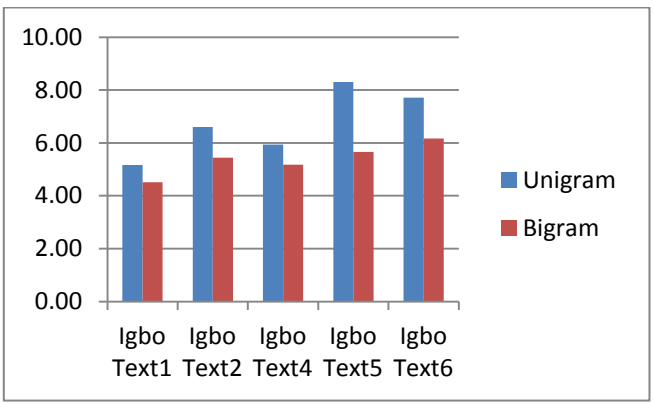

**Fig 7: Similarity Average Score on Unigram and Bigram Represented Igbo Text Chart**

## 5. RESULT ANALYSIS

Table 3 and Table 4 show the results obtained in computing the similarity measure on the same Igbo texts represented in unigram and bigram respectively. The average similarity value is obtained in each of the text representation model, summarized and displayed in Table 5. The results displayed in Table 3, Table 4 and Table 5 are also represented in Fig. 5, Fig. 6, and Fig. 7 respectively with chart diagrams.

Considering the results obtained in the first rows of Table 3 and Table 4 for the result analysis. In Table 3 representing result obtained in unigram represented text, the distance between Igbo text1 and Igbo textDoc1 is 6.78; the distance between Igbo text1 and Igbo textDoc2 is 4.36; the distance between Igbo text1 and Igbo textDoc3 is 6.40; the distance between Igbo text1 and Igbo textDoc4 is 5.48; the distance between Igbo text1 and Igbo textDoc5 is 7.07; and the distance between Igbo text1 and Igbo textDoc6 is 7.28. Similarity is measured by the distance between the documents; the more the distance, the less the similarity and vice versa. The distance between Igbo text1 and Igbo textDoc2 (4.36) is the least, meaning that Igbo text1 is most similar to Igbo textDoc2 using unigram represented text in computing the similarity.

In Table 4, representing result obtained in bigram represented text, the distance between Igbo text1 and Igbo textDoc1 is 5.00; the distance between Igbo text1 and Igbo textDoc2 is 6.00; the distance between Igbo text1 and Igbo textDoc3 is 4.47; the distance between Igbo text1 and Igbo textDoc4 is 3.61; the distance between Igbo text1 and Igbo textDoc5 is 7.48; and the distance between Igbo text1 and Igbo textDoc6 is 5.00. The distance between Igbo text1 and Igbo textDoc4 (3.61) is the least value, signifying that Igbo text1 is most similar to Igbo textDoc4 using bigram represented text. Ideally, Igbo text1 is most similar to Igbo textDoc4, which is the accurate result obtained with bigram represented text.

Table 5 shows the average distances of Igbo text1, Igbo text2, Igbo text4 and Igbo text5 obtained on unigram and bigram represented texts. The result shows that unigram represented text has highest distance values and bigram has the lowest corresponding distance values. The lower the distance value, the more similar the two documents and better the quality of the model, when used for a task that requires similarity measure. The similarity of two documents increases as the distance value moves down to zero (0).

## 6. CONCLUSION

The advancement of Information Technology has necessitated the invention and use of Igbo language in the creation of resources, articles and news online. As the number of Igbo text online and tasks that require computing similarity between Igbo text documents are increasing, this paper has presented a comparative analysis of n-gram text representation on Igbo text document similarity measure. The result analyzed shows that Igbo text document similarity computed on bigram represented text gives perfect accurate similarity result. This indicates that using bigram text representation on tasks that require computing Igbo text document similarity will give better and effective result. This is because Igbo text document is compounding in nature and is not duly captured with a unigram text representation model.

## 7. ACKNOWLEDGEMENTS

The authors wish to express their great appreciation to the reviewers of this paper. We are grateful of your useful comments and contributions that assisted in improving the value of this work.